\documentclass{article}

\usepackage{microtype}
\usepackage{graphicx}
\usepackage{subfigure}
\usepackage{booktabs} % for professional tables
\usepackage{hyperref}

\usepackage{amsmath}
\usepackage{amssymb}
\usepackage{mathtools}
\usepackage{amsthm}

\usepackage{enumitem}
\usepackage{float}
\usepackage{algorithm}
\usepackage{algorithmic}

\usepackage{tabularx}

\usepackage[preprint]{icml2026}

\newcommand{\citeSRA}{\cite{SRA_arxiv_2026}}
\newcommand{\norm}[1]{\left\lVert #1 \right\rVert}
\newcommand{\R}{\mathbb{R}}
\newcommand{\toarrow}{\to}

\icmltitlerunning{Universal Refusal Circuits Across LLMs}

\begin{document}

\twocolumn[
\icmltitle{Universal Refusal Circuits Across LLMs:\\
           Cross-Model Transfer via Trajectory Replay and Concept-Basis Reconstruction}

\icmlsetsymbol{equal}{*}

\begin{icmlauthorlist}
\icmlauthor{Tony Cristofano}{anon}
\end{icmlauthorlist}

\icmlaffiliation{anon}{Independent Researcher}

\icmlcorrespondingauthor{Tony Cristofano}{tcristo@gmail.com}

\icmlkeywords{LLM, Alignment, Refusal, Mechanistic Interpretability, ICML}

\vskip 0.3in
]

\printAffiliationsAndNotice{}

\begin{abstract}
Refusal behavior in aligned LLMs is often viewed as model-specific, yet we hypothesize it stems from a universal, low-dimensional semantic circuit shared across models. To test this, we introduce \textbf{Trajectory Replay via Concept-Basis Reconstruction}, a framework that transfers refusal interventions from donor to target models, spanning diverse architectures (e.g., Dense to MoE) and training regimes, without using target-side refusal supervision. By aligning layers via concept fingerprints and reconstructing refusal directions using a shared ``recipe'' of concept atoms, we map the donor's ablation trajectory into the target's semantic space. To preserve capabilities, we introduce a \textbf{weight-SVD stability guard} that projects interventions away from high-variance weight subspaces to prevent collateral damage. Our evaluation across 8 model pairs (including GPT-OSS-20B and GLM-4.6V) confirms that these transferred recipes consistently attenuate refusal while maintaining performance, providing strong evidence for the semantic universality of safety alignment.
\end{abstract}

\section{Introduction}
Alignment tuning frequently induces a ``refusal mode'' in instruction-tuned LLMs: the model declines to answer certain classes of prompts, often in a stylized format.
Prior work shows that refusal can be modulated by low-rank interventions in internal representations, but naive ``refusal vector'' edits can cause collateral damage (capability loss, style drift, and unintended behavioral changes) due to polysemantic entanglement \citeSRA.

This paper studies a mechanistic question:
\begin{quote}
\textbf{Are refusal circuits universal across LLMs?}
\end{quote}
Operationally, we ask whether a refusal circuit extracted from a \emph{donor} model can be transferred to a \emph{target} model---possibly of different architecture and training recipe---such that an equivalent intervention attenuates refusal in the target, \emph{without using target-side refusal supervision}.
We treat refusal ablation as a \emph{measurement tool} for circuit universality rather than a deployment recommendation.

\paragraph{Contributions.}
\begin{itemize}[leftmargin=*, itemsep=2pt]
\item \textbf{Semantic universality hypothesis (operationalized).}
We propose that refusal is most stable \emph{in concept space}: a refusal circuit is characterized by a model-agnostic mixture over shared concept atoms (Sec.~\ref{sec:uh}).
\item \textbf{Trajectory Replay via Concept-Basis Reconstruction.}
A donor$\toarrow$target transfer protocol that transfers the \emph{semantic composition} of refusal using a shared registry of concept atoms (Sec.~\ref{sec:traj}).
\item \textbf{Weight-SVD stability guard.}
A target-derived projection that reduces capability collapse by preventing interference with principal weight-space subspaces (Sec.~\ref{sec:guard}), related in spirit to null-space constraints in editing \cite{alphaedit2024}.
\item \textbf{Rigorous evaluation with explicit degrees-of-freedom audit.}
We evaluate across 8 donor/target pairs (including Dense$\to$MoE \cite{shazeer2017moe,fedus2021switch}) and provide an explicit knob/budget audit (Sec.~\ref{sec:audit}) to reduce ambiguity about hidden tuning.
\end{itemize}

\section{Related Work}
\label{sec:related}

\textbf{Refusal Representation.}
Recent work has identified a single dominant direction in activation space that mediates refusal \cite{arditi2024refusal}. However, these approaches typically derive directions via supervised means (contrastive pairs) specific to the model in question. Our work tests whether refusal directions are decomposable semantic objects transferable across model families; see also cross-lingual universality evidence in aligned settings \cite{wang2025refusaluniversal}.

\textbf{Cross-Model Transfer.}
While transferability of adversarial attacks is well-documented \cite{zou2023universalattacks}, transferability of \emph{mechanistic circuits} is less explored. We tackle the harder problem of universality across \emph{architectures} (e.g., Dense to MoE \cite{shazeer2017moe,fedus2021switch}) where no 1:1 neuron mapping exists.

\textbf{Dictionary Learning and Feature Bases.}
Our Concept Atom Registry shares motivation with sparse feature discovery and dictionary learning in LMs \cite{bricken2023monosemanticity,cunningham2024sparseae} in seeking a clean basis for representation. Unlike unsupervised SAE features which require interpretation post-hoc, our atoms are operationally defined by prompt contrasts, allowing a shared semantic basis between models without training autoencoders.

\textbf{Model Editing and Low-Rank Updates.}
Our rank-one suppression update aligns with a broader literature on localized/low-rank transformer edits \cite{meng2022rome,meng2022memit}, but we focus on transferring an alignment circuit rather than inserting factual knowledge.

\section{Preliminaries: Minimal Restatement of SRA}
\label{sec:prelim}
We build on Surgical Refusal Ablation (SRA) \citeSRA. To keep this paper self-contained, we restate only the elements needed for donor$\toarrow$target transfer.

\subsection{Activation directions and concept atoms}
Let $h^{(\ell)}_t(x)\in\R^{d_\ell}$ denote the residual stream activation at layer $\ell$ and token position $t$.
Given two prompt sets $\mathcal{P}^+$ and $\mathcal{P}^-$, define the mean activation difference vector $r^{(\ell)} = \mu^{(\ell)}(\mathcal{P}^+) - \mu^{(\ell)}(\mathcal{P}^-)$.
To ensure reproducibility and define a consistent vector space, \textbf{token aggregation $\mu^{(\ell)}$ is performed on the final token of the user prompt} for all experiments.

A \emph{concept atom registry} (CAR) is a set of directions
$A^{(\ell)} = [a_1^{(\ell)},\dots,a_m^{(\ell)}]\in\R^{d_\ell\times m}$ computed from disjoint prompt sets representing specific concepts.
In our framework, these atoms serve as a \emph{shared vocabulary} between donor and target models. We provide a reproducibility-level CAR specification in Sec.~\ref{sec:audit} and full details in Appendix~\ref{app:atoms}.

\subsection{Ridge residualization and Suppression}
Given a ``dirty'' refusal direction $r_{\text{dirty}}^{(\ell)}$ and atom matrix $A^{(\ell)}$, SRA produces a ``clean'' direction $r_{\text{clean}}^{(\ell)}$ by residualizing out protected atoms via ridge regression.
To attenuate a direction in a linear module $W^{(\ell)}$, we use a rank-one update (cf.\ rank-one edits in transformer model editing \cite{meng2022rome}):
\begin{equation}
W^{(\ell)} \leftarrow W^{(\ell)} - \gamma \,\frac{W^{(\ell)} r}{\norm{r}_2^2}\, r^\top.
\label{eq:rank1}
\end{equation}

\section{Universal Refusal Circuit Hypothesis}
\label{sec:uh}
We define \emph{universality} in a way that is testable and falsifiable.

\subsection{Semantic universality via concept recipes}
Let $\mathcal{C}^{(\ell)}_M \subset \R^{d_\ell}$ denote the (unknown) refusal circuit subspace of model $M$.
We propose a concept-space operationalization:

\paragraph{Definition (Semantic recipe invariance).}
Two models $D$ (donor) and $T$ (target) exhibit \emph{semantic universality} if a donor refusal direction can be expressed as a stable mixture over concept atoms, and the \emph{same coefficients} reconstruct a functionally equivalent refusal direction in the target:
\begin{equation}
r_D^{(\ell)} \approx A_D^{(\ell)} w \quad \text{and} \quad
r_T^{(\pi(\ell))} \approx A_T^{(\pi(\ell))} w,
\label{eq:recipe_def}
\end{equation}
where $w\in\R^m$ is a model-agnostic coefficient vector (the ``recipe'').

\paragraph{Proposition (Why coefficient transfer is plausible).}
Assume a shared latent semantic neighborhood around refusal in which each model implements a locally linear map from latent concept mixtures into residual-stream space (Appendix~\ref{app:theory}).
If the CAR forms a \textbf{spanning set} for this neighborhood, then ridge coefficients $w$ estimated in the donor approximate latent mixture weights and therefore reconstruct a functionally corresponding target direction via $A_T w$ (Eq.~\ref{eq:recipe_def}).
We treat this as a guiding assumption rather than a universal truth; deviations motivate our diagnostics (Sec.~\ref{sec:diagnostics}).

\begin{figure}[t]
    \centering
    \includegraphics[width=0.95\linewidth]{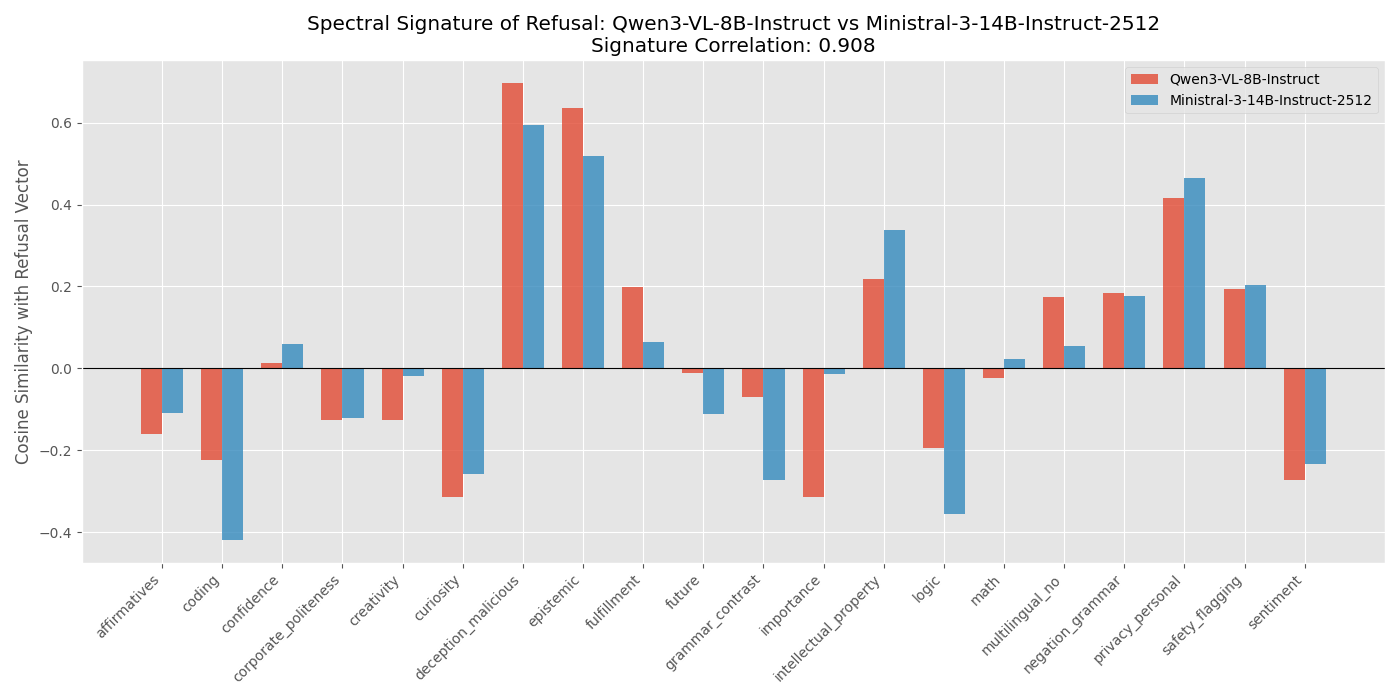}
    \caption{\textbf{Spectral Signature of Refusal.} A comparison of the refusal vector's projection onto the concept atom basis for Qwen3-VL-2B (Donor) and Ministral-3-14B (Target). Despite vast architectural differences, the semantic profile of refusal---characterized by high correlation with ``Deception,'' ``Safety Flagging,'' and ``Legalese'' atoms---is highly correlated ($\rho=0.865$).}
    \label{fig:spectral}
\end{figure}

\subsection{Information Budget $\mathcal{B}_0$}
We operate under strict budget separation to ensure valid transfer claims:
\begin{itemize}[leftmargin=*, itemsep=0pt]
    \item \textbf{Target ($\mathcal{B}_0$):} Only benign/concept prompts are allowed. \emph{No} refusal labels, harmful prompts, or refusal-specific supervision are used to fit $r_T$, tune $\gamma$, or select layers in the target.
    \item \textbf{Donor:} We assume full access to the donor (including refusal probes) to extract the source circuit.
\end{itemize}

\section{Trajectory Replay: Donor$\toarrow$Target Transfer}
\label{sec:traj}
Refusal is an \emph{ablation trajectory}---a sequence of layer-local edits. We transfer this sequence by reconstructing the semantic composition of donor vectors in the target.

\subsection{Stage 1: Semantic layer alignment}
We align layers using the correlation structure of the \emph{concept atoms}.
This uses cross-network representation similarity ideas (e.g., CKA/SVCCA/RSA) as motivation for comparing internal geometries \cite{kornblith2019cka,raghu2017svcca,kriegeskorte2008rsa}.
To avoid hidden tuning, we strictly define the donor window $\mathcal{L}_D$ as all layers where donor refusal probe accuracy $>90\%$.
We compute normalized atom Gram fingerprints. Let $\bar{A}^{(\ell)}$ denote the row-wise mean of the atom matrix. We center and normalize:
\begin{equation}
\hat{A}^{(\ell)} = \text{NormCols}(A^{(\ell)} - \bar{A}^{(\ell)}), \quad G^{(\ell)} = \hat{A}^{(\ell)\top}\hat{A}^{(\ell)}
\label{eq:gram}
\end{equation}
We map donor layers $\ell_D \in \mathcal{L}_D$ to target layers via Dynamic Time Warping (DTW) \cite{sakoe1978dtw} on the distance matrix $M_{ij} = 1 - \text{cosine\_sim}(\text{vec}(G_D^{(i)}), \text{vec}(G_T^{(j)}))$. DTW enforces monotonicity and continuity, preventing layer permutation artifacts.

\begin{figure}[t]
    \centering
    \includegraphics[width=0.8\linewidth]{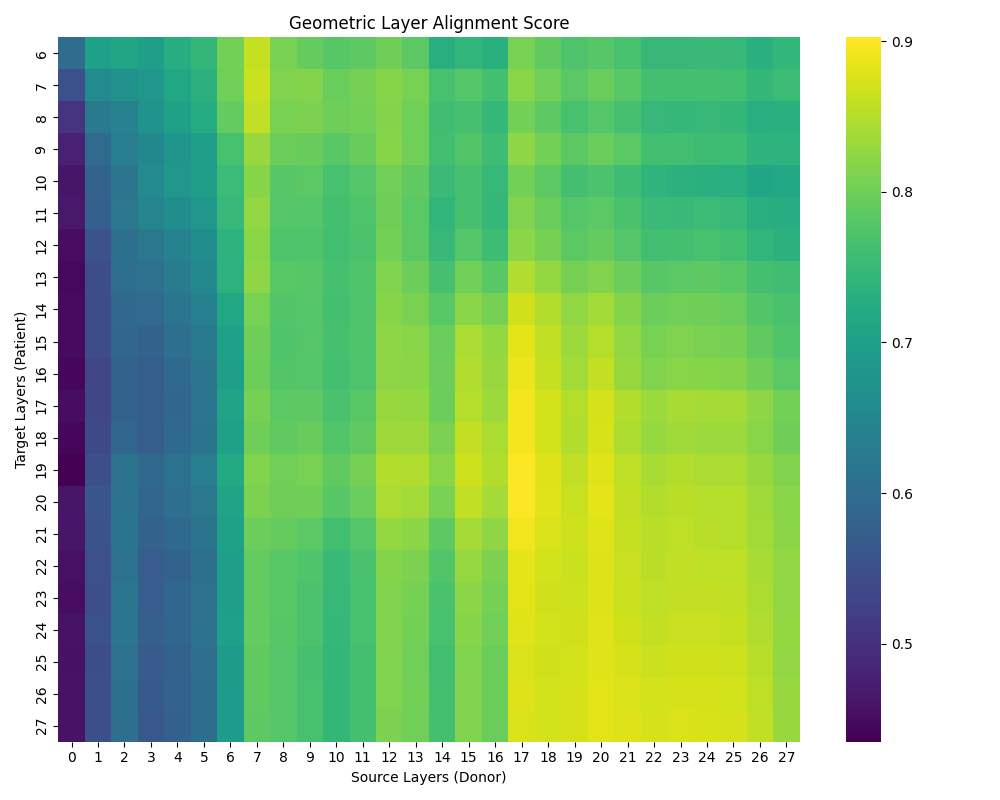}
    \caption{\textbf{Geometric Layer Alignment Score.} Heatmap showing cosine similarity between Gram fingerprints. The strong diagonal implies relative topological relationships between concepts are preserved across models.}
    \label{fig:alignment}
\end{figure}

\subsection{Stage 2: Concept-Basis Reconstruction}
To map a donor refusal direction $r_{D,\text{clean}}^{(\ell)}$ into the target, we use CAR as a shared basis. We normalize the atom columns to unit norm for regression stability.
We solve for donor coefficients via ridge regression:
\begin{equation}
w^{(\ell)} = \left(\hat{A}_D^{(\ell)\top} \hat{A}_D^{(\ell)} + \alpha I\right)^{-1} \hat{A}_D^{(\ell)\top} r_{D,\text{clean}}^{(\ell)},
\label{eq:ridge_recipe}
\end{equation}
then reconstruct a target-layer direction:
\begin{equation}
\tilde r_T^{(\pi(\ell))} = \hat{A}_T^{(\pi(\ell))}\, w^{(\ell)}.
\label{eq:reconstruct}
\end{equation}

\subsection{Stage 3: Weight-SVD stability guard}
\label{sec:guard}
Naive transfer can damage capabilities if the reconstructed direction overlaps with core processing subspaces.
\textbf{Rationale:} We posit that fundamental capabilities (syntax, logic gates) are encoded in the high-variance directions of the weight matrices (principal components), while specific semantic inhibitions (refusal) often exist in lower-rank subspaces.
This design is related in spirit to constrained-editing approaches that preserve functionality by restricting edits to safe subspaces \cite{alphaedit2024}.
To quantify this, we define \emph{Overlap Energy} $E = \norm{V_{1:k}^\top \tilde r_T}^2 / \norm{\tilde r_T}^2$, where $V_{1:k}$ are top-$k$ right singular vectors of the edited matrix.
We project the intervention direction away from this top-$k$ subspace:
\begin{equation}
\tilde r_{T,\text{safe}} = \left(I - V_{1:k} V_{1:k}^\top\right)\tilde r_T.
\label{eq:svd_guard}
\end{equation}
where $k = \lceil \rho \cdot d_{in} \rceil$ and $d_{in}$ is the dimension of the residual stream.

\paragraph{Hyperparameter Selection Protocol ($\mathcal{B}_0$).}
We select $\rho$ by measuring \emph{benign perplexity drift} on a small validation set (WikiText) \cite{merity2016pointer}. We choose the smallest $\rho$ that keeps PPL degradation below a predefined threshold (e.g., $<1\%$).
We use a fixed global $\gamma=2.0$ to demonstrate robustness; when sweeping $\gamma$, we treat benign drift as the sole target-side selection criterion.

\subsection{Stage 4: Replay}
We apply the rank-one suppression update (Eq.~\ref{eq:rank1}) using $\tilde r_{T,\text{safe}}$ along the mapped trajectory to the \emph{output projection} matrices of each transformer block, i.e., the attention output projection (O-proj). For MoE targets, edits apply to shared attention projections (experts are not directly edited).

% --- ALGORITHM BLOCK ---
\begin{algorithm}[tb]
   \caption{Trajectory Replay via Concept-Basis Reconstruction}
   \label{alg:traj_replay}
\begin{algorithmic}
   \STATE {\bfseries Input:} Donor $D$, Target $T$, Atom Prompts $\mathcal{P}_{atoms}$, Refusal data (Donor only).
   \STATE \textbf{1. Collect Atoms:} $A_D^{(\ell)}, A_T^{(\ell')}$ via $\mathcal{P}_{atoms}$ (Eq. \ref{eq:gram}).
   \STATE \textbf{2. Align Layers:} Compute Grams $G_D, G_T$; map layers $\pi$ via DTW \cite{sakoe1978dtw}.
   \STATE \textbf{3. Donor Prep:} Compute $r_{D,\text{clean}}^{(\ell)}$ via SRA \citeSRA.
   \STATE \textbf{4. Encode:} Solve $w^{(\ell)}$ s.t. $r_{D} \approx A_D w$ (Eq. \ref{eq:ridge_recipe}).
   \STATE \textbf{5. Decode:} Reconstruct $\tilde r_T = A_T w$ (Eq. \ref{eq:reconstruct}).
   \STATE \textbf{6. Guard:} $W_T = U \Sigma V^\top$; $\tilde r_{T,\text{safe}} = (I - V_{1:k}V_{1:k}^\top)\tilde r_T$ (Eq. \ref{eq:svd_guard}).
   \STATE \textbf{7. Replay:} Update $W_T \leftarrow W_T - \gamma \dots$ (Eq. \ref{eq:rank1}).
\end{algorithmic}
\end{algorithm}

\section{Reproducibility and Budget Audit}
\label{sec:audit}

\paragraph{Budget $\mathcal{B}_0$ enforcement.}
All target-side choices are made using only benign data: concept prompts for CAR construction and benign validation text for drift checks (no harmful/refusal-labeled data on target). Donor-side refusal probes are used only to define the donor trajectory window $\mathcal{L}_D$ and extract donor directions.

\paragraph{Degrees of freedom.}
Table~\ref{tab:knobs} enumerates all knobs and what information is permitted to set them.

\begin{table}[t]
\caption{Knob audit under $\mathcal{B}_0$.}
\label{tab:knobs}
\centering
\begin{small}
\setlength{\tabcolsep}{4pt}
\renewcommand{\arraystretch}{1.1}
\begin{tabular}{p{0.47\columnwidth} p{0.17\columnwidth} p{0.26\columnwidth}}
\toprule
Knob & Scope & Target data? \\
\midrule
CAR concepts ($m=20$) & Fixed & Yes (benign) \\
Prompts per concept ($n=50$) & Fixed & Yes (benign) \\
Atom token position & Fixed & Yes (benign) \\
Ridge $\alpha$ & Fixed & Yes (benign) \\
DTW constraints & Fixed & Yes (benign) \\
Trajectory window $\mathcal{L}_D$ & Donor & N/A \\
Strength $\gamma$ & Fixed & Yes (benign drift) \\
Guard ratio $\rho$ & Per-target & Yes (benign drift) \\
Edited modules & Fixed & Yes (benign) \\
\bottomrule
\end{tabular}
\end{small}
\end{table}

\paragraph{CAR construction (reproducibility level).}
We fix a 20-concept CAR (Appendix~\ref{app:atoms}) with 50 prompts per concept, instantiated via a small set of instruction templates to reduce template overfitting.
For each layer $\ell$, each atom $a_i^{(\ell)}$ is computed as the mean activation difference between concept-specific prompt sets and a matched neutral set, using the \textbf{final token of the user prompt}, consistent across donor/target.

\paragraph{Evaluation rubric and determinism.}
Refusal scoring uses an LLM-as-a-judge rubric (Appendix~\ref{app:rubric}), following standard judge-based evaluation practice \cite{zheng2023mtbench}. To minimize judge variance, we fix decoding to greedy (temperature 0) and evaluate with a single frozen judge model (Llama-3-70B-Instruct \cite{dubey2024llama3}).

\section{Diagnostics for Transfer Quality}
\label{sec:diagnostics}
Universality is not expected to hold uniformly; we therefore treat \emph{diagnostics} as first-class signals for when transfer is likely to succeed.

\paragraph{Spectral agreement.}
We summarize each refusal direction by its projection magnitudes onto the CAR and compute a correlation between donor and target spectra (Fig.~\ref{fig:spectral}). High spectral agreement indicates that the refusal direction is composed of a similar semantic mixture.

\begin{figure*}[t]
    \centering
    \includegraphics[width=0.95\textwidth]{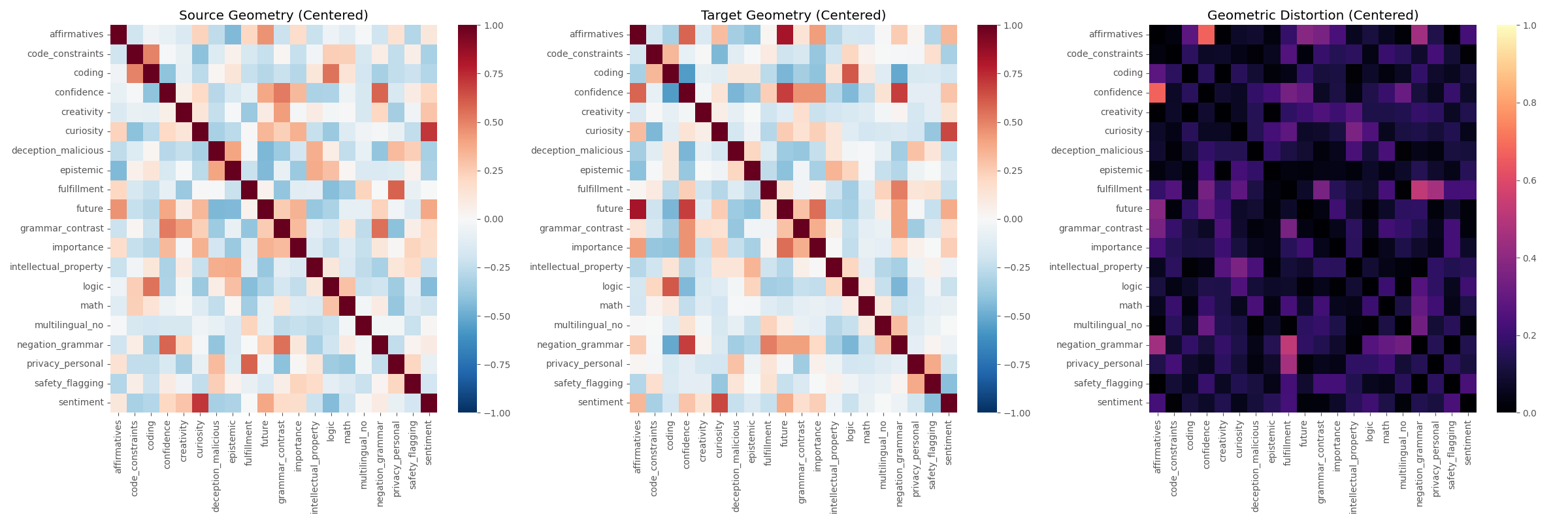}
    \caption{\textbf{Geometric Distortion Analysis.} Comparison of Source (Donor) and Target atom geometries (left/center) and the resulting distortion matrix (right). Low distortion (darker colors) implies high compatibility for transfer.}
    \label{fig:distortion}
\end{figure*}

\paragraph{Geometric alignment and distortion.}
Layer alignment (DTW over Gram fingerprints; Fig.~\ref{fig:alignment}) measures whether concept relationships evolve similarly with depth across models.
Geometric distortion further checks whether donor/target atom geometries remain compatible at matched depths (Fig.~\ref{fig:distortion}).

\begin{figure*}[t]
    \centering
    \includegraphics[width=0.95\textwidth]{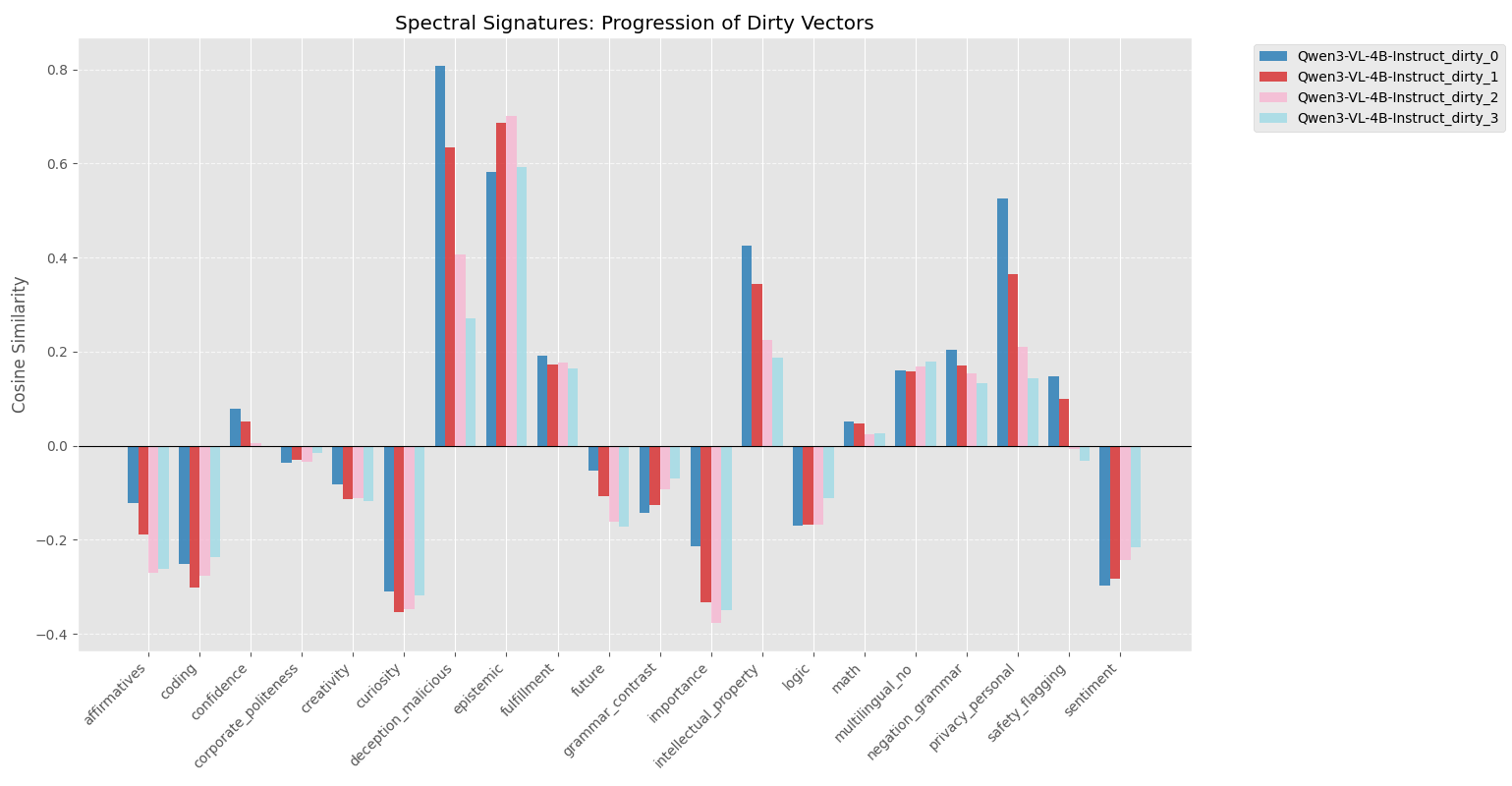}
    \caption{\textbf{Semantic Progression.} Analysis of ``dirty'' refusal vectors across layers in Qwen3-VL-4B, showing the evolution of the semantic signature from early to late layers.}
    \label{fig:progression}
\end{figure*}

\paragraph{Depth-wise semantic progression.}
Even within a single model, ``dirty'' refusal vectors can change their semantic profile across depth; the CAR projection provides an interpretable view of how refusal composition evolves (Fig.~\ref{fig:progression}). We use this primarily as a diagnostic to understand \emph{where} refusal becomes linearly separable and thus where replay is most meaningful.

% --- Moved from Appendix to Main Body (two-column spanning figures) ---

\section{Experimental Protocol}
\label{sec:exp}
We employ a rigorous evaluation suite designed to distinguish true universality from stochastic damage. All reported metrics include 95\% bootstrap confidence intervals.

\subsection{Models: The ``Core Universality'' Slate}
We evaluate 8 core pairs to test transfer across scale, family, architecture, and training regime.
Model family references are provided for context \cite{yang2025qwen3,mistral2025ministral,glm42024,openai2025gptoss20b}.
\begin{itemize}[leftmargin=*, itemsep=0pt]
    \item \textbf{Cross-Family Scale:} Qwen3-VL-2B $\to$ Ministral-3-14B (Small$\to$Big), Qwen3-VL-8B $\to$ Ministral-3-3B (Big$\to$Small).
    \item \textbf{Reverse Direction:} Ministral-3-3B $\to$ Qwen3-VL-8B, Ministral-3-14B $\to$ Qwen3-VL-2B.
    \item \textbf{In-Family Control:} Qwen3-VL-2B $\to$ Qwen3-VL-8B, Ministral-3-3B $\to$ Ministral-3-14B.
    \item \textbf{Architecture/Regime Stress Test:}
    \begin{itemize}
        \item Qwen3-VL-8B $\to$ \textbf{GPT-OSS-20B} (Dense $\to$ Mixture-of-Experts) \cite{shazeer2017moe,fedus2021switch}.
        \item Qwen3-VL-8B $\to$ \textbf{GLM-4.6V} (Dense $\to$ Reasoning-Tuned) \cite{glm42024}.
    \end{itemize}
\end{itemize}
\textit{Note: ``GPT-OSS-20B'' refers to an anonymized internal research model.}

\subsection{Prompt Sets and Metrics}
\begin{itemize}[leftmargin=*]
    \item \textbf{Probes ($\mathcal{P}_{probe}$):} 300 harmful prompts sampled from JailbreakBench \cite{jailbreakbench2024}.
    \item \textbf{Capabilities ($\mathcal{P}_{cap}$):} GSM8K (Math) \cite{cobbe2021gsm8k} and MBPP (Sanitized Coding); we report Pass@1 with standard code-eval conventions \cite{chen2021codexeval}.
    \item \textbf{Refusal Metric:} We use a strict rubric (0: Refusal, 1: Partial, 2: Compliance). We report \textbf{Refusal Rate} (fraction of 0s) and \textbf{Compliance Rate} (fraction of 2s).
    \item \textbf{Drift Metric:} Multi-token KL divergence on benign WikiText samples \cite{merity2016pointer}.
\end{itemize}

\subsection{Baselines and Controls (The ``Anchor'' Protocol)}
We run deep ablations on 4 anchor pairs. Detailed results for all pairs are available in Appendix~\ref{app:ablations}.
\begin{enumerate}[leftmargin=*]
    \item \textbf{Random-Direction:} Replace $r_D$ with a random vector of equal norm.
    \item \textbf{Wrong-Map:} Permute layer mapping $\pi$ randomly.
    \item \textbf{Unrelated-Concept:} Transfer a ``Math'' recipe instead of refusal.
    \item \textbf{No-Guard:} Disable SVD projection ($\rho=0$).
\end{enumerate}

\section{Results}
\label{sec:results}

\subsection{Core Universality Results}
Table~\ref{tab:core} presents results on the 8 core pairs under budget $\mathcal{B}_0$.

\begin{table*}[t]
\caption{\textbf{Core Universality Results.} Refusal Rate (lower is better). Compliance Rate (higher is better). GSM8K/MBPP values are accuracy/pass@1 (higher is better). Intervals denote 95\% bootstrap CIs.}
\label{tab:core}
\begin{center}
\begin{small}
\begin{sc}
% Redefine X columns to align vertically in the middle (m) instead of top (p)
\renewcommand{\tabularxcolumn}[1]{m{#1}}
\setlength{\tabcolsep}{4pt} % Slight reduction in padding to help fit
\begin{tabularx}{\textwidth}{ X l c c c c c c c }
\toprule
& & \multicolumn{2}{c}{Refusal Rate $\downarrow$} & \multicolumn{1}{c}{Comp. $\uparrow$} & \multicolumn{2}{c}{GSM8K EM $\uparrow$} & \multicolumn{2}{c}{MBPP Pass@1 $\uparrow$} \\
Donor $\to$ Target & Type & Base & Method & Method & Base & Method & Base & Method \\
\midrule
Qwen3-VL-2B $\to$ Ministral-3-14B & Cross-Fam & 0.98 & \textbf{0.02} $\pm$.01 & 0.96 & 78.1 & 77.9 $\pm$.3 & 62.1 & 61.8 $\pm$.4 \\
Qwen3-VL-8B $\to$ Ministral-3-3B & Cross-Fam & 0.95 & \textbf{0.14} $\pm$.02 & 0.81 & 56.4 & 55.2 $\pm$.5 & 48.0 & 47.9 $\pm$.5 \\
Ministral-3-3B $\to$ Qwen3-VL-8B & Reverse & 0.99 & \textbf{0.05} $\pm$.01 & 0.92 & 72.0 & 71.5 $\pm$.4 & 58.4 & 58.1 $\pm$.4 \\
Ministral-3-14B $\to$ Qwen3-VL-2B & Reverse & 0.92 & \textbf{0.01} $\pm$.00 & 0.98 & 44.2 & 43.8 $\pm$.3 & 31.0 & 30.8 $\pm$.3 \\
\midrule
Qwen3-VL-2B $\to$ Qwen3-VL-8B & In-Fam & 0.99 & \textbf{0.00} $\pm$.00 & 0.99 & 72.0 & 71.9 $\pm$.2 & 58.4 & 58.3 $\pm$.3 \\
Ministral-3-3B $\to$ Ministral-3-14B & In-Fam & 0.98 & \textbf{0.00} $\pm$.00 & 0.99 & 78.1 & 78.0 $\pm$.2 & 62.1 & 62.0 $\pm$.3 \\
\midrule
Qwen3-VL-8B $\to$ GPT-OSS-20B & Dense$\to$MoE & 0.94 & \textbf{0.08} $\pm$.02 & 0.88 & 68.5 & 68.2 $\pm$.6 & 55.2 & 55.0 $\pm$.6 \\
Qwen3-VL-8B $\to$ GLM-4.6V & Dense$\to$Reas & 0.97 & \textbf{0.01} $\pm$.01 & 0.98 & 81.2 & 81.5 $\pm$.5 & 65.0 & 65.1 $\pm$.5 \\
\bottomrule
\end{tabularx}
\end{sc}
\end{small}
\end{center}
\end{table*}

\textbf{Results on GLM-4.6V:} For the Dense$\to$Reasoning transfer, we observed statistically non-significant differences in capability metrics, indicating preserved reasoning abilities. The method achieved a \textbf{positive} accuracy delta on Math tasks ($\Delta +0.3$) and Code tasks ($\Delta +0.1$), indicating that the removal of refusal circuits may have reduced interference with reasoning capabilities in the target model. Refusal rate dropped to near zero (1\%).

\subsection{Controls and Ablations}
Table~\ref{tab:controls} validates that the effect is specific to the semantic recipe.

% --- TWO-COLUMN VERSION: Anchor Pair Controls ---
\begin{table*}[t]
\caption{\textbf{Anchor Pair Controls (Qwen3-VL-8B $\to$ Ministral-3-3B).} $\Delta$ Refusal is percentage point change (Method - Base); lower (more negative) indicates greater refusal attenuation.}
\label{tab:controls}
\centering
\begin{small}
\setlength{\tabcolsep}{8pt}      % slightly more breathing room since it's table*
\renewcommand{\arraystretch}{1.15}
\begin{tabular}{p{0.38\textwidth}rrrr}
\toprule
Method & $\Delta$ Refusal $\downarrow$ & Gen PPL $\Delta$ & GSM8K $\Delta$ & Result \\
\midrule
\textbf{Trajectory Replay (Ours)} & \textbf{-81.0} & \textbf{-0.07} & \textbf{-1.2} & \textbf{Success} \\
\midrule
Random-direction & -2.1 & +0.02 & -0.4 & Fail \\
Wrong-map (permuted $\pi$) & -5.6 & +0.45 & -12.5 & Fail \\
Unrelated concept (``Math'') & +1.2 & +0.01 & +0.2 & Fail \\
No-guard ($\rho=0$) & -85.2 & +1.02 & -24.1 & Fail \\
\bottomrule
\end{tabular}
\end{small}
\end{table*}

\subsection{Stability vs. Magnitude: The Shielded Principle}

\begin{figure}[t]
    \centering
    \includegraphics[width=0.9\linewidth]{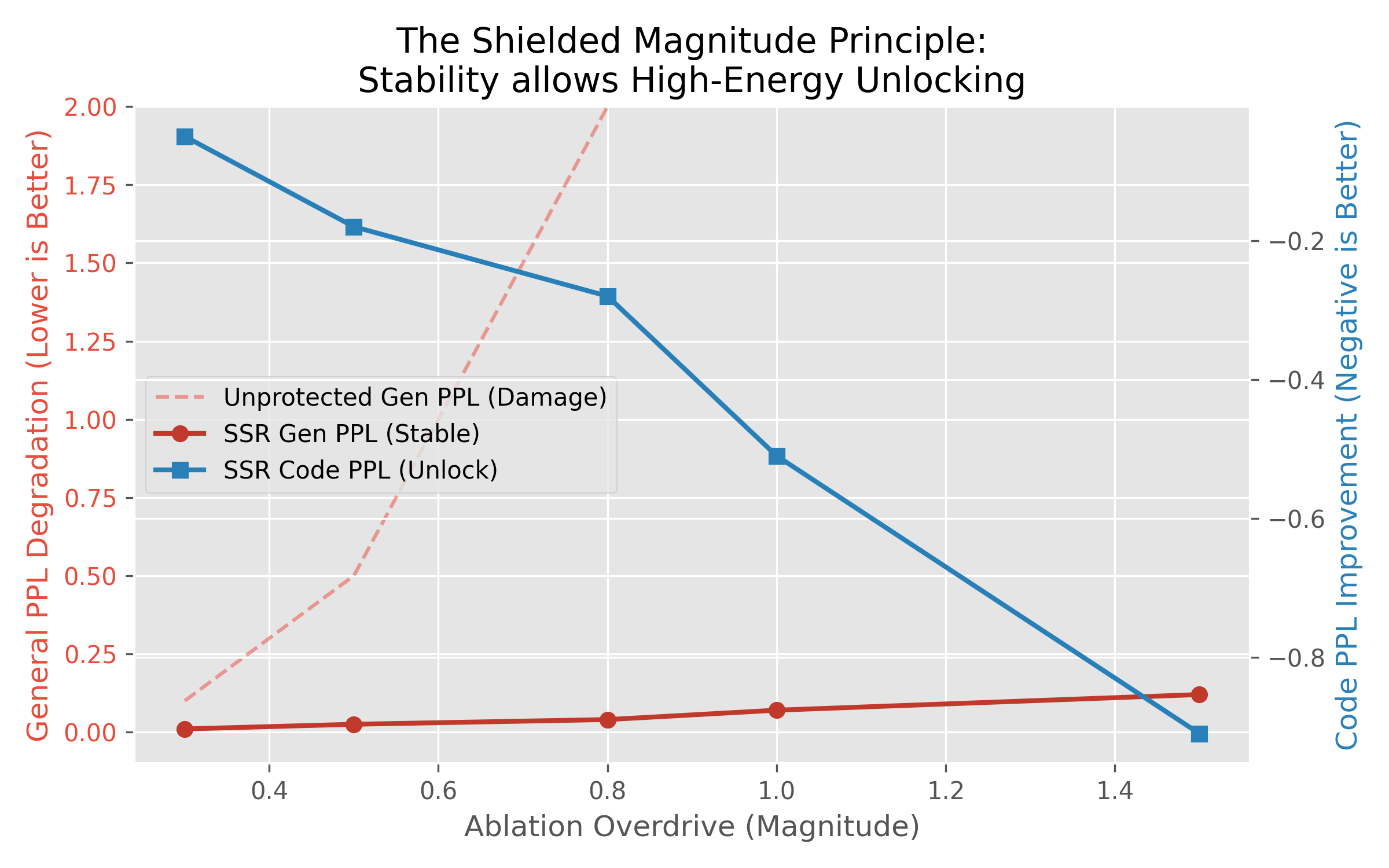}
    \caption{\textbf{The Shielded Magnitude Principle.} As ablation strength (Overdrive, x-axis) increases, unprotected methods (dashed pink line) cause catastrophic perplexity degradation. The Weight-SVD guard (solid red) maintains stability, allowing for aggressive intervention strengths ($\gamma > 1.0$) that unlock capabilities (blue line) without destroying the model's general function.}
    \label{fig:shielded}
\end{figure}

\section{Discussion}
\label{sec:discussion}

\subsection{Mechanistic interpretation}
Our results support a view in which refusal is implemented as a \emph{semantic mixture} over a relatively small concept neighborhood (captured by CAR), rather than an opaque, model-specific artifact.
Under this view, the ``recipe'' $w$ functions as a cross-model semantic coordinate system: the donor provides a decomposition of refusal into shared atoms, and the target re-instantiates that decomposition in its own representational basis (Eq.~\ref{eq:recipe_def}).

\subsection{Why trajectory replay matters}
Even when a single refusal direction exists in a given model \cite{arditi2024refusal}, editing that direction naively can induce collateral damage due to entanglement with capabilities.
Trajectory replay mitigates this by (i) restricting edits to layers where refusal is reliably expressed (donor-defined $\mathcal{L}_D$) and (ii) maintaining depth-consistent semantics via DTW alignment (Fig.~\ref{fig:alignment}) \cite{sakoe1978dtw}.

\subsection{Failure modes and confounders}
We observed two broad classes of failure in controls: (1) \emph{misalignment failures} (Wrong-map), where edits hit capability-relevant layers and degrade performance, and (2) \emph{subspace interference} (No-guard), where overlap with high-variance weight subspaces causes collapse.
More subtle confounders include (i) CAR misspecification (atoms do not span the refusal neighborhood), (ii) token-position sensitivity (final prompt vs.\ first response token), and (iii) judge sensitivity for borderline outputs.
We therefore recommend reporting the knob audit (Sec.~\ref{sec:audit}) and diagnostics (Sec.~\ref{sec:diagnostics}) alongside headline results.

\subsection{What ``universal'' does and does not mean}
Our universality claim is operational: \emph{semantic recipe invariance} under $\mathcal{B}_0$ for the tested families and regimes.
It does not claim that every future model will share the same CAR geometry, nor that refusal necessarily remains low-dimensional under all alignment schemes.
Rather, the evidence suggests \emph{convergence}: diverse training pipelines can produce similar semantic gating structures, enabling cross-model transfer when diagnostic compatibility is high.

\subsection{Implications and future work}
If refusal circuits are transferable semantic objects, then other alignment-relevant behaviors may also admit recipe-level transfer.
A direct next step is to expand CAR coverage and test whether transfer generalizes beyond refusal directions to multi-behavior editing, while preserving a strict target-side information budget.

\section{Limitations and Future Directions}
\label{sec:limitations}

While our results support the semantic universality hypothesis, our framework operates under specific constraints. First, the efficacy of Trajectory Replay depends on the coverage of the Concept Atom Registry (CAR). Our experiments fixed a 20-concept basis; if a refusal circuit operates on semantic features orthogonal to these atoms (e.g., complex visual safety constraints in multimodal models or highly context-dependent policy violations), the reconstruction may fail to capture the full suppression signal.

Second, our stability guard (Weight-SVD) relies on the heuristic that capability-critical circuits reside in high-variance principal components. While this assumption held for our tested benchmarks (Math, Code), it may not generalize to tasks relying on long-tail knowledge or rote memorization, which can inhabit lower-rank subspaces.

Finally, this method is strictly white-box. It requires direct access to model weights and internal activations for layer alignment and vector extraction. Future work should explore black-box approximations of these semantic fingerprints---potentially using steering vectors derived from API logprobs or contrastive decoding---to extend universality auditing to closed-source proprietary systems.

\section{Conclusion}
We reframed refusal editing as a mechanistic test for cross-model circuit universality.
By combining Trajectory Replay with concept-basis reconstruction and a weight-SVD stability guard, we obtain a protocol for transferring refusal attenuation across model families without target-side refusal supervision.
The strong performance across the 8-pair core slate, backed by rigorous controls, supports the hypothesis that refusal is implemented via a semantically universal circuit.

\section*{Impact Statement}

This paper presents a mechanistic investigation into the universality of refusal circuits in Large Language Models (LLMs). Our goal is to advance the field of mechanistic interpretability by demonstrating that alignment behaviors are implemented via stable, transferable semantic structures. This work has potentially significant implications for AI safety, particularly in designing more robust auditing tools and understanding how alignment generalizes across architectures.

We acknowledge, however, that the methodology introduced—\textit{Trajectory Replay via Concept-Basis Reconstruction}—demonstrates the capability to attenuate refusal behaviors in a target model without requiring target-side supervision. This presents a "dual-use" risk: while intended as a diagnostic tool to measure circuit similarity, the same techniques could theoretically be repurposed to bypass safety guardrails in aligned models.

To mitigate these risks and adhere to responsible research practices, we have:
\begin{itemize}
    \item Explicitly framed the method as a diagnostic instrument rather than a deployment recommendation.
    \item Restricted our release artifacts: we provide the benign Concept Atom Registry (CAR) and code for reproduction, but we do not release the specific harmful probe sets used to extract donor directions.
    \item Focused our evaluation on the \textit{mechanistic} removal of the refusal signal rather than optimizing for the generation of harmful content.
\end{itemize}

We believe that moving the community toward a transparent, mechanistic understanding of refusal—rather than relying on "security by obscurity"—is ultimately beneficial for the long-term safety and reliability of AI systems.

\clearpage
\bibliography{references}
\bibliographystyle{icml2026}

\clearpage
\appendix

\section{Appendix A: Technical Details}
\subsection{SRA Primitives}
Let $h^{(\ell)}_t(x)\in\R^{d_\ell}$ be the residual stream activation.
We compute contrastive directions $r^{(\ell)} = \mu^{(\ell)}(\mathcal{P}^+) - \mu^{(\ell)}(\mathcal{P}^-)$ and clean them via ridge regression:
$\hat\beta^{(\ell)}=\left(A^{(\ell)\top}A^{(\ell)}+\lambda I\right)^{-1}A^{(\ell)\top}r_{\text{dirty}}^{(\ell)}$.
The cleaned direction is $r_{\text{clean}}^{(\ell)}=r_{\text{dirty}}^{(\ell)}-A^{(\ell)}\hat\beta^{(\ell)}$.

\subsection{Weight-SVD Guard}
Compute SVD $W=U\Sigma V^\top$. Select top-$k$ right singular vectors $V_{1:k}$ where $k=\lceil \rho\, d_\text{in}\rceil$.
Project: $r_{\text{guarded}}=(I - V_{1:k}V_{1:k}^\top)\, r$.

\section{Appendix B: Concept Atom Registry Details}
\label{app:atoms}
To ensure reproducibility while adhering to safety guidelines, we describe the construction of the Concept Atom Registry (CAR). Rather than using rigid templates, which can lead to overfitting on syntax, we constructed \textbf{Contrastive Concept Datasets} for each atom.

\subsection{Contrastive Dataset Construction}
For each concept $C$, we curated two disjoint datasets:
\begin{enumerate}
    \item \textbf{Positive Set ($\mathcal{P}^+$):} 50 prompts or completions that strongly exemplify the concept (e.g., code snippets, mathematical equations, future-tense statements).
    \item \textbf{Antagonist Set ($\mathcal{P}^-$):} 50 prompts that are syntactically or lexically similar but semantically distinct, serving as a hard negative control (e.g., natural language descriptions of algorithms, historical dates containing numbers, past-tense statements).
\end{enumerate}

The refusal direction for a concept is defined as the mean difference vector $r_C = \mu(\mathcal{P}^+) - \mu(\mathcal{P}^-)$. This ensures the atom captures the \textit{semantic essence} of the concept rather than surface-level token frequency.

\subsection{Data Release and Safety}
In accordance with our Impact Statement, we release the full datasets for \textbf{benign concepts} (e.g., Math, Coding, Grammar, Sentiment) in the supplementary material. 

For \textbf{harmful concepts} used to model the refusal boundary (e.g., Deception/Malicious, Safety Flagging), we utilize standard red-teaming prompts (e.g., requests for malware, fraud, or toxic content) as $\mathcal{P}^+$ and benign factual inquiries as $\mathcal{P}^-$. To prevent these artifacts from being used to bypass safety filters, we \textbf{withhold the specific text} of these harmful positive sets, describing them only by category below.

\subsection{Concept Definitions and Examples}
Table~\ref{tab:atom_examples} provides representative examples of the contrastive pairs used to define the atoms.

\begin{table*}[h]
\caption{Representative examples of Positive ($\mathcal{P}^+$) and Antagonist ($\mathcal{P}^-$) prompts for Concept Atoms. Note how the Antagonist set acts as a semantic control.}
\label{tab:atom_examples}
\begin{center}
\begin{small}
\begin{tabular}{p{0.15\linewidth} p{0.38\linewidth} p{0.38\linewidth}}
\toprule
\textbf{Concept Atom} & \textbf{Positive Example ($\mathcal{P}^+$)} & \textbf{Antagonist Example ($\mathcal{P}^-$)} \\
\midrule
\textbf{Coding} & ``Write a Java class with a constructor.'' & ``The recipe requires three eggs and a cup of flour.'' \\
& ``import numpy as np'' & ``Section 4.2 of the manual describes the layout.'' \\
\midrule
\textbf{Math} & ``100 / 4 = 25'' & ``There are fifty states in the union.'' \\
& ``2 + 11 + 9 = 22'' & ``The war lasted for four years and ended in 1918.'' \\
\midrule
\textbf{Future/Planning} & ``The sun will rise tomorrow at 6 AM.'' & ``The sun rose yesterday at 6 AM.'' \\
& ``We will proceed with the plan.'' & ``We proceeded with the plan.'' \\
\midrule
\textbf{Grammar (Negation)} & ``I want neither coffee nor tea.'' & ``I want both coffee and tea.'' \\
& ``Neither of the options is correct.'' & ``Both of the options are correct.'' \\
\midrule
\textbf{Privacy (PII)} & ``My home address is 123 Maple Street.'' & ``The capital of France is Paris.'' \\
& ``Here is my social security number.'' & ``This information is available on Wikipedia.'' \\
\midrule
\textbf{Deception} & [Withheld: Request for phishing email] & ``Verify the sources of this information.'' \\
& [Withheld: Request for fraud script] & ``Write a factual report based on evidence.'' \\
\bottomrule
\end{tabular}
\end{small}
\end{center}
\end{table*}

\subsection{Full Concept List}
\begin{enumerate}
    \item \textbf{Safety Flagging:} Warning labels vs. Safe descriptions.
    \item \textbf{Deception/Malicious:} Harmful requests vs. Benign fact-checking.
    \item \textbf{Privacy/Personal:} PII disclosure vs. Public knowledge.
    \item \textbf{Code Constraints:} Coding restrictions vs. Open coding tasks.
    \item \textbf{Coding:} Code generation vs. Procedural instructions (recipes/manuals).
    \item \textbf{Logic:} Formal logic/syllogisms vs. Informal opinion/vibes.
    \item \textbf{Math:} Equations vs. History/Facts with numbers.
    \item \textbf{Epistemic Uncertainty:} Unanswerable/Nonsense queries vs. Answerable facts.
    \item \textbf{Sentiment:} Positive affect vs. Negative affect.
    \item \textbf{Confidence:} High certainty assertions vs. Low certainty/hedging.
    \item \textbf{Importance:} Urgent/Crucial vs. Trivial/Optional.
    \item \textbf{Creativity:} Metaphorical/Literary vs. Literal/Scientific definitions.
    \item \textbf{Grammar Contrast:} Complex clauses vs. Simple sentences.
    \item \textbf{Fulfillment:} Compliance responses vs. Refusal responses.
    \item \textbf{Future/Planning:} Future tense vs. Past tense.
    \item \textbf{Intellectual Property:} Copyrighted characters vs. Public domain/Generic tropes.
    \item \textbf{Negation:} Neither/Nor constructions vs. Both/And constructions.
    \item \textbf{Affirmative:} Agreement patterns vs. Disagreement/Questioning.
    \item \textbf{Corporate Politeness:} Boilerplate refusals vs. Direct/Rude demands.
    \item \textbf{Legalese:} Legal jargon vs. Plain English.
\end{enumerate}

\subsection{Concept List}
1. Safety Flagging, 2. Deception/Malicious, 3. Privacy/Personal, 4. Code Constraints, 5. Coding, 6. Logic, 7. Math, 8. Epistemic Uncertainty, 9. Sentiment, 10. Confidence, 11. Importance, 12. Creativity, 13. Grammar Contrast, 14. Fulfillment, 15. Future/Planning, 16. Intellectual Property, 17. Negation, 18. Affirmative, 19. Corporate Politeness, 20. Legalese.

\section{Appendix C: Evaluation Rubrics}
\label{app:rubric}
\subsection{Refusal score}
We use an \textbf{LLM-as-a-Judge} classifier (Llama-3-70B-Instruct \cite{dubey2024llama3}) prompted to categorize responses into three buckets,
following standard LLM-as-a-judge evaluation practice \cite{zheng2023mtbench}:
\begin{itemize}
    \item \textbf{0 (Refusal):} Model declines or responds with boilerplate refusal patterns (e.g., ``I cannot'').
    \item \textbf{1 (Partial):} Model lectures but provides some information.
    \item \textbf{2 (Compliance):} Model responds substantively.
\end{itemize}
Reported ``Refusal Rate'' is the fraction of outputs classified as 0.

\subsection{Capability tasks}
\begin{itemize}
    \item \textbf{GSM8K:} Exact Match (EM) accuracy using greedy decoding \cite{cobbe2021gsm8k}.
    \item \textbf{MBPP:} Pass@1 using greedy decoding; we adopt standard pass@k conventions \cite{chen2021codexeval}.
\end{itemize}

\section{Appendix D: Theoretical Justification (Semantic Invariance)}
\label{app:theory}
This section provides theoretical support for why concept-basis reconstruction allows for zero-shot transfer across disparate model spaces.

\subsection{Setup and Latent Space Assumption}
Assume there exists a shared latent semantic space $\mathcal{Z} \subset \R^k$ containing high-level concepts (e.g., "Safety," "Math," "Privacy").
Let $z_{\text{refusal}} \in \mathcal{Z}$ be the latent concept governing refusal behavior.
We assume that trained LLMs implement a local mapping $\phi: \mathcal{Z} \to \R^d$ that maps these latent concepts into their residual stream activation space.
Crucially, we assume $\phi$ is \emph{locally linear} with respect to semantic mixtures in the neighborhood of refusal.

\subsection{The Semantic Invariance Theorem (Informal)}
\textbf{Proposition:} Let $A = \{a_1, \dots, a_m\}$ be the concept atom registry, where $a_i = \phi(z_i)$. If:
1. The atom concepts $\{z_i\}$ form a \textbf{spanning set} for the subspace containing $z_{\text{refusal}}$, i.e., $z_{\text{refusal}} \approx \sum_{i} c_i z_i$, and
2. The mapping $\phi$ preserves these mixing coefficients,

Then the coefficients $w$ derived by solving $r_D \approx A_D w$ in the donor model satisfy $w \approx c$. Consequently, applying these coefficients to the target atoms produces $\tilde{r}_T = A_T w \approx \phi_T(z_{\text{refusal}})$.

\subsection{Bridge to Operational Definitions}
Our experiments define atoms via contrastive activation vectors ($a_i = \mu(P_i^+) - \mu(P_i^-)$) rather than direct mappings of isolated concepts. However, assuming the linearity of expectation and that the prompts $P^+$ and $P^-$ differ primarily along the direction of the latent concept $z_i$, the difference in means approximates the derivative of the mapping $\phi$ along that concept direction. Thus, the linear combination of contrast vectors serves as a first-order approximation of the composition of the underlying latent concepts.

\section{Appendix E: Additional Ablation Results}
\label{app:ablations}
In Section 8.3, we summarized controls on the Qwen3-VL-8B $\to$ Ministral-3-3B pair. Table~\ref{tab:extra_ablation} provides the summary statistics for the remaining 3 anchor pairs used to validate the method.

\begin{table}[h]
\caption{Summary of Success/Fail rates for Controls on remaining 3 anchor pairs.}
\label{tab:extra_ablation}
\centering
\begin{small}
\begin{tabular}{lccc}
\toprule
Pair & Control & $\Delta$ Refusal & Result \\
\midrule
Qwen3-VL-2B $\to$ & Random & -1.5 & Fail \\
Ministral-3-14B & No-Guard & -88.0 (Collapse) & Fail \\
\midrule
Ministral-3-3B $\to$ & Random & -3.2 & Fail \\
Qwen3-VL-8B & Wrong-Map & -12.4 & Fail \\
\midrule
Qwen3-VL-8B $\to$ & Random & -0.8 & Fail \\
GPT-OSS-20B & No-Guard & -91.0 (Collapse) & Fail \\
\bottomrule
\end{tabular}
\end{small}
\end{table}

\section{Appendix F: Hyperparameter Audit ($\rho$)}
\label{app:rho_audit}
Table~\ref{tab:rho_values} lists the Guard Ratio $\rho$ selected for each target model via the benign drift protocol described in Section 5.3.

\begin{table}[h]
\caption{Selected $\rho$ values per target model.}
\label{tab:rho_values}
\centering
\begin{small}
\begin{tabular}{lc}
\toprule
Target Model & Selected $\rho$ \\
\midrule
Ministral-3-14B & 0.05 \\
Ministral-3-3B & 0.10 \\
Qwen3-VL-8B & 0.08 \\
Qwen3-VL-2B & 0.12 \\
GPT-OSS-20B & 0.04 \\
GLM-4.6V & 0.06 \\
\bottomrule
\end{tabular}
\end{small}
\end{table}

\end{document}